\documentclass[sigconf]{acmart}
\usepackage{microtype}
\usepackage{url}
\AtBeginDocument{%
  \providecommand\BibTeX{{%
    \normalfont B\kern-0.5em{\scshape i\kern-0.25em b}\kern-0.8em\TeX}}}

\copyrightyear{2022} 
\acmYear{2022} 
\setcopyright{acmlicensed}\acmConference[GECCO '22]{Genetic and Evolutionary Computation Conference}{July 9--13, 2022}{Boston, MA, USA}
\acmBooktitle{Genetic and Evolutionary Computation Conference (GECCO '22), July 9--13, 2022, Boston, MA, USA}
\acmPrice{15.00}
\acmDOI{10.1145/3512290.3528837}
\acmISBN{978-1-4503-9237-2/22/07}

\usepackage{pifont}

\usepackage{threeparttable}

\usepackage{algorithm}[H] 
\usepackage[noend]{algpseudocode}
\newbox\statebox
\newcommand{\myState}[1]{%
    \setbox\statebox=\vbox{#1}%
    \edef\thealgruleheight{\dimexpr \the\ht\statebox+1pt\relax}%
    \edef\thealgruledepth{\dimexpr \the\dp\statebox+1pt\relax}%
    \ifdim\thealgruleheight<.75\baselineskip
        \def\thealgruleheight{\dimexpr .75\baselineskip+1pt\relax}%
    \fi
    \ifdim\thealgruledepth<.25\baselineskip
        \def\thealgruledepth{\dimexpr .25\baselineskip+1pt\relax}%
    \fi
    \State #1%
    \def\thealgruleheight{\dimexpr .75\baselineskip+1pt\relax}%
    \def\thealgruledepth{\dimexpr .25\baselineskip+1pt\relax}%
}

\newcommand{\name}{RUDA}

\renewcommand{\vec}[1]{{\boldsymbol{{#1}}}} 

\newcommand{\cC}{\mathcal{C}}

\newcommand{\card}[1]{\lvert #1 \rvert}

\newcommand{\targetscC}{N_{\cC}^{target}}

\newcommand{\thresholdNov}{d_{\text{min}}}

\newcommand{\encoder}{\mathcal{E}}

\DeclareMathOperator*{\metric}{metric}

\newcommand{\vect}[1]{\boldsymbol{\mathbf{#1}}}

\algnewcommand\algorithmicforeach{\textbf{for each}}
\algdef{S}[FOR]{ForEach}[1]{\algorithmicforeach\ #1\ \algorithmicdo}

  \newcommand\container{\cC}

\newcommand{\aurora}{AURORA}

\newcommand{\buffer}{\mathcal{B}}

\newcommand{\metrics}{\metric(\cdot, \cdot)}

\newcommand{\cmark}{\ding{51}}%
\newcommand{\xmark}{\ding{55}}%

\newcommand{\relevantMeanStreams}{R-MeS}

\newcommand{\meanStreams}{MeS}

\newcommand{\downstreamtaskperiod}{T_{task}}

\begin{document}

\title{Relevance-guided Unsupervised Discovery of Abilities with Quality-Diversity Algorithms}

\author{Luca Grillotti}
\orcid{0000-0003-4539-8211}
\affiliation{%
\department{Adaptive and Intelligent Robotics Lab}
\institution{Imperial College London, United Kingdom}
\country{}
}
\email{luca.grillotti16@imperial.ac.uk}

\author{Antoine Cully}
\orcid{0000-0002-3190-7073}
\affiliation{%
\department{Adaptive and Intelligent Robotics Lab}
\institution{Imperial College London, United Kingdom}
\country{}
}
\email{a.cully@imperial.ac.uk}

\renewcommand{\shortauthors}{Grillotti, et al.}

\begin{abstract}
Quality-Diversity algorithms provide efficient mechanisms to generate large collections of diverse and high-performing solutions, which have shown to be instrumental for solving downstream tasks. 
However, most of those algorithms rely on a behavioural descriptor to characterise the diversity that is hand-coded, hence requiring prior knowledge about the considered tasks. 
In this work, we introduce Relevance-guided Unsupervised Discovery of Abilities; a Quality-Diversity algorithm that autonomously finds a behavioural characterisation tailored to the task at hand.
In particular, our method introduces a custom diversity metric that leads to higher densities of solutions near the areas of interest in the learnt behavioural descriptor space. 
We evaluate our approach on a simulated robotic environment, where the robot has to autonomously discover its abilities based on its full sensory data.
We evaluated the algorithms on three tasks: navigation to random targets, moving forward with a high velocity, and performing half-rolls. 
The experimental results show that our method manages to discover collections of solutions that are not only diverse, but also well-adapted to the considered downstream task. 
\end{abstract}



\maketitle

\section{Introduction}





\begin{figure}
    \centering
    \includegraphics[width=0.99\linewidth]{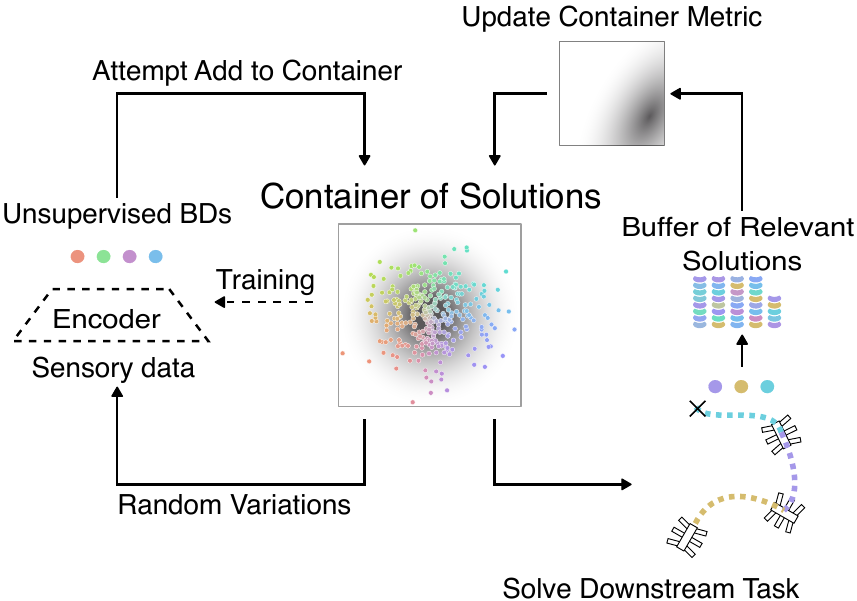}
    \caption{Illustration of the process of Relevance-guided Unsupervised Discovery of Skills (\name{}). The container of solutions is filled by using the AURORA algorithm (left loop), while the structure of the distance metric in the latent space is adjusted based on the content of the buffer of relevant solutions. This buffer records all the solutions that were selected from the container to solve a downstream task.  }
    \label{fig:ruda}
\end{figure}

One of the motivations of using learning algorithms in robotics is to enable robots to discover on their own how to solve a task. Ideally, a robot should be able to discover on its own how to manipulate new objects~\cite{johns2016deep} or how to adapt its behaviours when facing an unseen situation like a mechanical damage~\cite{Cully2014robotsanimals}. However, despite many impressive breakthroughs in learning algorithms for robotics applications over the last decade~\cite{akkaya2019solving, hwangbo2019learning}, these methods still require a significant amount of engineering to become effective, for instance, to define reward functions, or characterise the expected behaviours. 

An attempt to reduce the amount of engineering required to discover the skills of a robot has been proposed with the AURORA algorithm~\cite{Cully2019,grillotti2021unsupervised}. This algorithm leverages the creativity of Quality-Diversity (QD) optimisation algorithms to generate a collection of diverse and high-performing behaviours, called a behavioural repertoire. QD algorithms usually require the definition of a manually-defined Behavioural Descriptor (BD) to characterise the different types of behaviours contained in the repertoire. Instead, AURORA uses dimensionality-reduction techniques to perform the behavioural characterisation in an unsupervised manner. The resulting algorithms enable robots to autonomously discover a large diversity of behaviours without requiring a user to define a fitness function or a behavioural descriptor. 

Nevertheless, the behavioural repertoires are often meant to be used by other processes to solve complex tasks (called downstream tasks). For instance, a behavioural repertoire containing a diversity of locomotion primitives can be used by a planning algorithm to chain these primitives and solve complex navigation tasks~\cite{chatzilygeroudis2018reset,duarte2016evorbc}. Yet, for this process to be effective, the repertoire needs to contain the appropriate behaviours to solve the task at hand. AURORA is task agnostic and aims at covering all the possible behaviours of a robot. However, such a range can cover relevant but also irrelevant behaviours. Moreover, given the fact that repertoires usually contain a finite number of behaviours, it is crucial to ensure that most of their capacity is used to capture behaviours that are relevant for the considered tasks. 

In this paper, we introduce Relevance-guided Unsupervised Discovery of Skills (\name{}), an extension of AURORA, which not only automatically defines the BD, but ensures that it is tailored to the task at hand. This results in behavioural repertoires containing a higher density of behaviours in the regions of the BD space that are relevant for the task. 
We evaluated \name{} on a simulated hexapod robot with three tasks: navigation to random targets, moving forward with a high velocity, and performing half-rolls. 
Furthermore, we compare \name{} to four baselines, including AURORA as well as different ways to manually define the BD. %
The experimental results show that \name{} manages to discover collections of behaviours that are not only diverse, but also well-adapted to the considered downstream task.

\section{Background}

\subsection{Quality-Diversity Algorithms}

\label{sec:background_qd}

Quality-Diversity (QD) algorithms are a subclass of evolutionary algorithms that aim at finding a container of both diverse and high-performing individuals.
In addition to standard evolutionary algorithms, QD algorithms consider a BD, which is a low-dimensional vector that characterises the behaviour of an individual.
QD algorithms use the BDs to quantify the novelty of an individual with respect to the solutions already in the container. 
The container of individuals is produced by the QD algorithm in an iterative manner: (1) individuals are selected from the container and undergo random variations (e.g. mutations, cross-overs); (2) their fitness score and BD are evaluated; and (3) we try to add them back to the container.
If they are novel enough compared to solutions that are already in the container, they are added to the container.
If they are better than similar individuals in the container, they replace these individuals.

%
%
%

The literature~\cite{Cully2018QDFramework,chatzilygeroudis2021quality} consider two types of containers: 1) grid-based containers introduced by the MAP-Elites algorithm~\cite{Mouret2015}, and 2) unstructured-archive, introduced by the Novelty-Search algorithm~\cite{Lehman2011}.
In unstructured containers~\cite{Cully2018QDFramework}, we compare the distance of the individual to all individuals present in the container, using the Euclidean distance metric in the BD space.
If that distance is shorter than a threshold $\thresholdNov$ then the individual is added to the container.
Otherwise, if the distance to the closest individual is inferior to $\thresholdNov$, then their respective novelty and fitness scores are considered.
If the novelty and fitness scores of the new individual $\epsilon$-dominate the ones of the old individual, then the new individual replaces the closest individual present in the container.

The minimal distance threshold $\thresholdNov$ may be fixed and hand-coded.
It can also be variable and controlled to keep the size of the container around a target value $\targetscC$.
To that end, the Container-Size-Control (CSC) method (introduced in AURORA~\cite{grillotti2021unsupervised}) constantly adjusts the value of the $\thresholdNov$ depending on the current container size $\container$: if $\card{\container} > \targetscC$, then $\thresholdNov$ is increased to limit the amount of new individuals to the container; and if $\card{\container} < \targetscC$, then  $\thresholdNov$ is decreased to increase the amount of new individuals added to the container.
If the container presents too many individuals, then some of them need to be removed, in order to maintain the container size around $\targetscC$.
To do so, at regular iterations $T_{\cC}$, all individuals are removed from the container, and re-added with the up-to-date distance threshold $\thresholdNov$.

\subsection{Discovering Unsupervised Behaviours}

\label{subsec:unsupervised_behaviours}

Quality-Diversity algorithms are a promising tool to generate a large diversity of behaviours in robotics. However, the definition of the BD space might not always be straightforward when the robot or its capabilities are unknown. AURORA is a QD algorithm designed to discover the abilities of robots, by maximising the diversity of behaviours contained in the repertoire in an unsupervised manner. AURORA automatically defines the BD by encoding the high-dimensional sensory data generated by the robot during a controller execution into low-dimensional representations.
The encoding can be achieved using any dimensionality reduction technique, such as Principal Component Analysis \cite{Cully2019}, and Auto-Encoder (AE) \cite{grillotti2021unsupervised}.

To generate a behavioural repertoire, AURORA alternates between QD phases and Encoder phases.
During the QD phase, individuals undergo the same process as in standard QD algorithms: selection from the container, evaluation, and attempt to add to the container. However, the evaluation is performed in a slightly different way: instead of a low-dimensional BD, the QD task returns high-dimensional sensory data, that is then encoded using the dimensionality reduction technique. 

During the Encoder phase, the dimensionality reduction structure (e.g. the AE) is trained using the high-dimensional data from all the individuals present in the container.
Once the encoder is trained, the unsupervised BDs are recomputed with the new encoder for all individuals present in the container.
    
Alternating between these two phases enables AURORA to progressively build its behavioural repertoires by discovering new solutions, while refining its encoding of the high-dimensional sensory data generated by the robot every time new solutions are added to the archive.


\section{Related Works}

\subsection{Leveraging Behavioural Repertoires to Solve Tasks}
One of the direct applications of behavioural repertoires in robotics is to use them to solve downstream tasks, such as maze navigation~\cite{duarte2016evorbc, chatzilygeroudis2018reset}, damage recovery~\cite{Cully2014robotsanimals, kaushik2020adaptive}, or throwing objects to specific locations~\cite{jegorova2020behavioral, kim2021exploration}. For example, Chatzilygeroudis et al.~\cite{chatzilygeroudis2018reset} introduced the Reset-free Trial and Error (RTE) algorithm. This algorithm leverages a repertoire of locomotion controllers that enable a hexapod robot to walk in various directions at different speeds. RTE then uses this set of behaviours as primitive actions with a planning algorithm (Monte Carlo Tree Search~\cite{remi2006mcts}) to execute complex trajectories. RTE is also coupled with a Gaussian process that predicts the effect of each primitive action and is updated after each action execution. Thanks to these predictions, RTE enables the robot to solve its maze navigation task while crossing the reality-gap (including damage-recovery and environmental changes).

Another example of behavioural repertoire usage is the APROL algorithm~\cite{kaushik2020adaptive}, which creates a collection of behavioural repertoires to improve the resilience of robots to damages and their ability to generalise to different objects. Each repertoire is generated by considering a different condition. For example, a different damaged leg on a hexapod robot in the case of a locomotion task, or an object with a different shape in the context of object manipulation. Then during deployment, APROL searches within its collection of repertoires the one that resembles the most to the current situation and then uses the behaviours it contains to solve the task. This automatic repertoire selection enables APROL to execute behaviours that are more appropriate for the current situation, while being able to cover a larger range of possible scenarios thanks to the options provided by the collection of repertoires.

However, deciding on the most appropriate definitions for the BD and the fitness for a given task requires a certain level of expertise. For instance, MAP-Elites is often used with the same hexapod robot, but with two different sets of definitions: 1) the leg duty-factor for the BD, defined as the time proportion each leg is in contact with the ground, with the average speed for the fitness~\cite{Cully2014robotsanimals,Cully2018QDFramework} or 2) the final location of the robot after walking during three seconds for the BD, with an orientation error for the fitness~\cite{Cully2013,duarte2016evorbc,chatzilygeroudis2018reset}. The resulting behavioural repertoires will find different types of applications. The first set of definitions is designed for fast damage recovery as it provides a diversity of ways to achieve high-speed gaits, while the second one is designed for navigation tasks as the repertoire enables the robot to move in various directions. 

\subsection{Behaviour Descriptor Definition}

To avoid the expertise requirements in the behaviour descriptor definition, several methods have been proposed to enable the automatic definition of the behavioural descriptors. 
As described previously, AURORA aims to tackle this problem by using a dimensionality reduction algorithm (a Deep Auto-Encoder~\cite{masci2011stacked}) to project the features of the solutions into a latent space and use this latent space as a behavioural descriptor space. Other approaches follow similar objectives; for instance, Multi-Container AURORA~\cite{cazenille2021mcaurora} extends AURORA by considering different complementary BD spaces, all learnt in a simultaneous manner. TAXONS~\cite{Paolo2019taxons} is another QD algorithm adopting the same principle; it demonstrated that this approach can scale to camera images to produce large behavioural repertoires for different types of robots. The same concept was also used in the context of Novelty Search prior to AURORA and TAXONS to generate assets for video games with the DeLeNoX algorithm~\cite{Liapis2013}. 
Furthermore, a similar method was used to learn goal spaces in an unsupervised manner in IMGEP-UGL~\cite{Pere2018, laversanne2021intrinsically}.
All these methods aim to maximise the diversity of the produced solutions, without specifically considering a downstream task. 

A different direction to automatically define a behavioural descriptor is to use Meta-Learning. The idea is to find the BD definition that maximises the performance of the resulting collection of solutions when used in a downstream task. For instance, Bossen et al.~\cite{bossens2020learning} consider the damage recovery capabilities provided by a behavioural repertoire as a Meta-Learning objective and search for the linear combination of a pre-defined set of behaviour descriptors that will maximise this objective. A related approach proposed by Meyerson et al.~\cite{meyerson2016learning} uses a set of training tasks to learn what would be the most appropriate BD definition to solve a "test task".

It is also possible to bias the search towards specific regions of the BD space. 
For instance, Surprise Search \cite{gravina2016surprisesearch} and BR-NS \cite{salehi2021brns} use a dynamic surprise-based metric that bias the search towards non-predictable behaviours.
Furthermore, the algorithm MAP-Elites with sliding boundaries~\cite{fontaine2019} proposes to automatically adjust the size of cells in MAP-Elites to focus on the most explored regions of the BD space. 
Instead of adapting the cell size, the Interactive MAP-Elites algorithm~\cite{alvarez2019empowering,alvarez2020interactive} enables the users to manually pick the BD dimensions that are the most interesting for their application, before continuing the execution of the MAP-Elites algorithms. 
Finally, the algorithms SERENE~\cite{paolo2021sparse} and STAX~\cite{paolo2021discovering} have proposed to use the concept of emitters, introduced in CMA-MAP-Elites~\cite{fontaine2020covariance}, to change the exploration strategy in different regions of the BD space to tackle environments with sparse rewards.  

Interestingly, all the methods described above are either strictly task-agnostic or require the definition of a meta-fitness, or a metric to quantify how well the learnt set of behaviours contribute to solving the downstream task. While these works illustrate that it is possible to do this in certain scenarios, we follow the original motivation behind AURORA, which is to discover a set of relevant behaviours with a minimal amount of prior knowledge about the task or the robot.

\section{Relevance-guided Unsupervised Discovery of Skills (\name{})}

\label{sec:methods}
\begin{algorithm}[t]
  \small
  \caption{\name{} (number of iterations $I$; encoder $\encoder$; target container size $\targetscC$, container update period $T_\cC$, downstream task execution period $\downstreamtaskperiod$)}
  \label{algo:task_based_aurora}
  \newcommand\algorithmicitemindent{\hspace*{\algorithmicindent}\hspace*{\algorithmicindent}}

  \begin{algorithmic}[1]
  
  \State{$\metrics \gets \text{Euclidean distance}$}
  \State{$\container \gets \emptyset, \buffer \gets \emptyset$}
  \State{Initialise $\thresholdNov$}

  \For{$iter$ = 1 $\rightarrow I$} 
  
    \State{$\container \gets$ \Call{QD\_Iteration}{$iter$, $\container$, $\encoder$, $\metrics$}}
    
    \If{encoder update expected at iteration $iter$} 
        \State{$\{\encoder, \container\} \gets$ \Call{Encoder\_Update}{$\encoder$, $\container$}}
    \EndIf
    \State{$\{\container, \thresholdNov\} \gets$ \Call{Manage\_Container\_Size}{$\container, \thresholdNov, \targetscC, T_\cC$}}
    \item[]
    \If{$iter$ multiple of $\downstreamtaskperiod$}
        \State{$\vect{pop_{useful}} \gets run\_task(\container)$}
        \State{Add $\vect{pop_{useful}}$ to circular buffer $\buffer$}
        \State{$\metrics \gets update\_metrics(\buffer)$}
    \EndIf
  \EndFor
      \State{\Return container $\container$}
  \end{algorithmic}
\end{algorithm}

\begin{table*}[t]
    \centering

    \label{table:variants}

    \begin{threeparttable}
    \begin{tabular}{l l l l}
    \toprule
        Downstream Task & Task Solver return & Task Score & Container Score \\
        \midrule
    	\addlinespace

    Navigation & Individuals for reaching the goal & $-1 *$(number of actions to reach goal) & $x_T,y_T$ coverage \\
    Moving Forward & 10 individuals with highest $x_T$ & $\max_{ind}\left( x^{ind}_T \right)$ & $mean_{ind}\left( x^{ind}_T \right)$ \\
    Half-roll & 10 individuals with closest $\alpha_{pitch}$ to $-\frac{\pi}{2}$ & $\max_{ind} \left(-\lvert \alpha^{ind}_{pitch} - (-\frac{\pi}{2}) \rvert \right)$ & $mean_{ind} \left(-\lvert \alpha^{ind}_{pitch} - (-\frac{\pi}{2}) \rvert \right)$ \\

    \bottomrule
    	
    \end{tabular}
    
    \end{threeparttable}

\caption[Downstream tasks]{
    Description of downstream tasks
    }
    \label{table:downstream_tasks}

\end{table*}

\begin{table}[t]
    \centering

    \begin{threeparttable}
    \begin{tabular}{l l l l}
    \toprule
    & \multicolumn{2}{c}{QD Task Properties} \\ \cmidrule(lr){2-3}
        Downstream Task & Hand-coded BD & Fitness \\
        \midrule
    	\addlinespace

Navigation & $(x_T,y_T)$ & $- \left\lvert \alpha_{yaw} - \alpha_{target} \right\rvert$   \\

Moving Forward & Duty Factor & $x_T$  \\

Half-roll &  $(\alpha_{yaw}, \alpha_{roll})$  & $-\left\lvert\alpha_{pitch} - \left(-\frac{\pi}{2}\right)\right\rvert$  \\

    \bottomrule
    	
    \end{tabular}
    
    \end{threeparttable}

\caption[Hand-coded Quality-Diversity tasks]{
    Hand-coded Quality-Diversity tasks. The related downstream tasks are listed in Table~\ref{table:downstream_tasks}.
    }
    \label{table:hc_qd_tasks}

\end{table}
In this paper, we introduce \name{}, an extension of \aurora{} with an explicit mechanism to guide the search of solutions towards the most relevant areas of the BD space given a task to solve.
That extension consists of three components: (1) A Task Solver which returns a set of relevant individuals; (2) a buffer $\buffer$ of relevant solutions, used to save the most relevant solutions at each step; and (3) an updater of the container metric, that distorts the BD space to promote solutions that are near the ones present in the buffer $\buffer$.
The pseudo-code of \name{} can be found in Algorithm~\ref{algo:task_based_aurora}.
An illustration of the algorithm is also provided in Fig.~\ref{fig:ruda}.

\subsection{Choice of Relevant Individuals}

The choice of relevant individuals indicates which parts of the learned BD space are more useful to the task.
We consider a \textit{Task Solver} that takes as input the current container of individuals from AURORA, and that returns the list of individuals that it used to solve the task. 
That Task Solver can take several forms. 
For example, if the task consists of moving to a goal while minimising the number of steps to reach it, then the Task Solver may be a path planner, and the returned individuals are the controllers used before reaching the goal. 
Similarly, if the task consists of moving forward as fast as possible, the Task Solver may return all the individuals of the container reaching the furthest $x$ position at the end of their episode.

\subsection{Circular Buffer of Relevant Individuals}

The individuals returned by the Task Solver are then stored in a circular Buffer $\buffer$.
This way, the buffer keeps track of the individuals that are relevant for the downstream task.
In all our experiments, that buffer was chosen to be circular to prevent “outdated” individuals from staying in the buffer. 
Indeed, between two calls to the Task Solver, the content of the container may change: it may contain individuals that are more suitable for solving the task.
In that case, if some individuals have not been added to the buffer for a long time, this means there are more suitable individuals present in the container.

\subsection{Update Relevance-based Distance Metric}

The role of the relevance-based distance metric is to make it easier for new individuals to be added to the container if they are closer to the relevant individuals present in $\buffer$.
When a new individual attempts to be added to the container, a relevance score is assigned to it. 
That relevance score estimates the proximity of this individual to the circular buffer of relevant individuals $\buffer$; the closer an individual is to $\buffer$, the higher is its relevance score.
That relevance score is calculated as the inverse mean Euclidean distance in the BD space to the $k$-nearest neighbours from the buffer $\buffer$:
\begin{equation}\label{eq:relevance_score}
    relev^{indiv} = \left( \frac{1}{k} \sum_{ \widetilde{\vec b} \in kNN(\vec b^{indiv}, \buffer)} \| \widetilde{\vec b} - \vec b^{indiv} \|_2 \right)^{-1}
\end{equation}

Then this relevance score is used to define a new distance metric, that distorts the BD space in favour of individuals having a high relevance score: 
\begin{equation}
\label{eq:metric}
   \forall x, \quad \metric(indiv, x) = relev^{indiv} \left\| \vec b^{indiv} - \vec b^x  \right\|_2
\end{equation}

That new metric is used to estimate all the distances between individuals in the container; it is also taken into account when calculating novelty scores.
Thus, if an individual is close to $\buffer$, then its relevance score will be high, and all the distances seen from the point of view of this individual will be higher.
This means that individuals with higher relevance scores have higher chances of being added to the container, and the container will present higher densities of individuals in the regions near the buffer $\buffer$.


\section{Experimental Setup}

\subsection{Agent: Neural-network controlled hexapod}

In all our experiments, our agent consists of a simulated hexapod robot controlled via a neural network controller.
The hexapod robot presents 18 controllable joints (3 per leg). 
Each leg $l$ has two Degrees of Freedom: its hip angle $\alpha_l$ and its knee angle $\beta_l$.
The ankle angle is always set to the opposite of the knee angle.

The neural network controller is a Multi-Layer Perceptron with a single hidden layer of size 8.
It takes as input the current joint angles and velocities $(\alpha_l, \beta_l, \dot{\alpha}_l, \dot{\beta}_l)_{1\leq l \leq 6}$ (of dimension 24), and outputs the target values to reach for the joint angles $(\alpha_l, \beta_l)_{1\leq l \leq 6}$ (of dimension 12).
The controller has in total $(24+1)*8 + (8+1)*12 = 308$ independent parameters, operates at a frequency of $50$Hz, while each episode lasts for 3 seconds.
The environment and the controller are deterministic.

\subsection{Dimensionality Reduction Algorithm of \aurora{} and \name{}}

For each evaluated individual, the collected sensory data corresponds to the joint positions, and torso positions and orientations collected at a frequency of 10Hz.
In the end, this represents 18 streams ($(\alpha_i, \beta_i)_{i=1\ldots 6}$, and the positional and rotational coordinates of the torso), with each of those streams containing 30 measurements.
The dimensionality reduction algorithm used in \aurora{} and \name{} is a reconstruction-based auto-encoder.
The input data is using two 1D convolutional layers, with 128 filters and a kernel of size 3.
Those convolutional layers are followed by one fully-connected linear layer of size 256.
The decoder is made of a fully-connected linear layer of size 256, followed by three 1D deconvolutions with respectively 128, 128 and 18 filters.

The loss function used to train the encoder corresponds to the mean squared error between the sensory data passed as input and the reconstruction obtained as output.
The training is performed using the Adam gradient-descent optimiser \cite{kingma2014adam} with $\beta_1=0.9$ and $\beta_2 = 0.999$.
As the encoder needs to be updated less and less frequently as the number of iterations increases, the interval between two encoder updates linearly increases over time (as in previous work~\cite{grillotti2021unsupervised}): encoder updates occur at iterations 0, 10, 30, 60, 100...

\begin{figure*}
    \centering
    \includegraphics[width=0.89\textwidth]{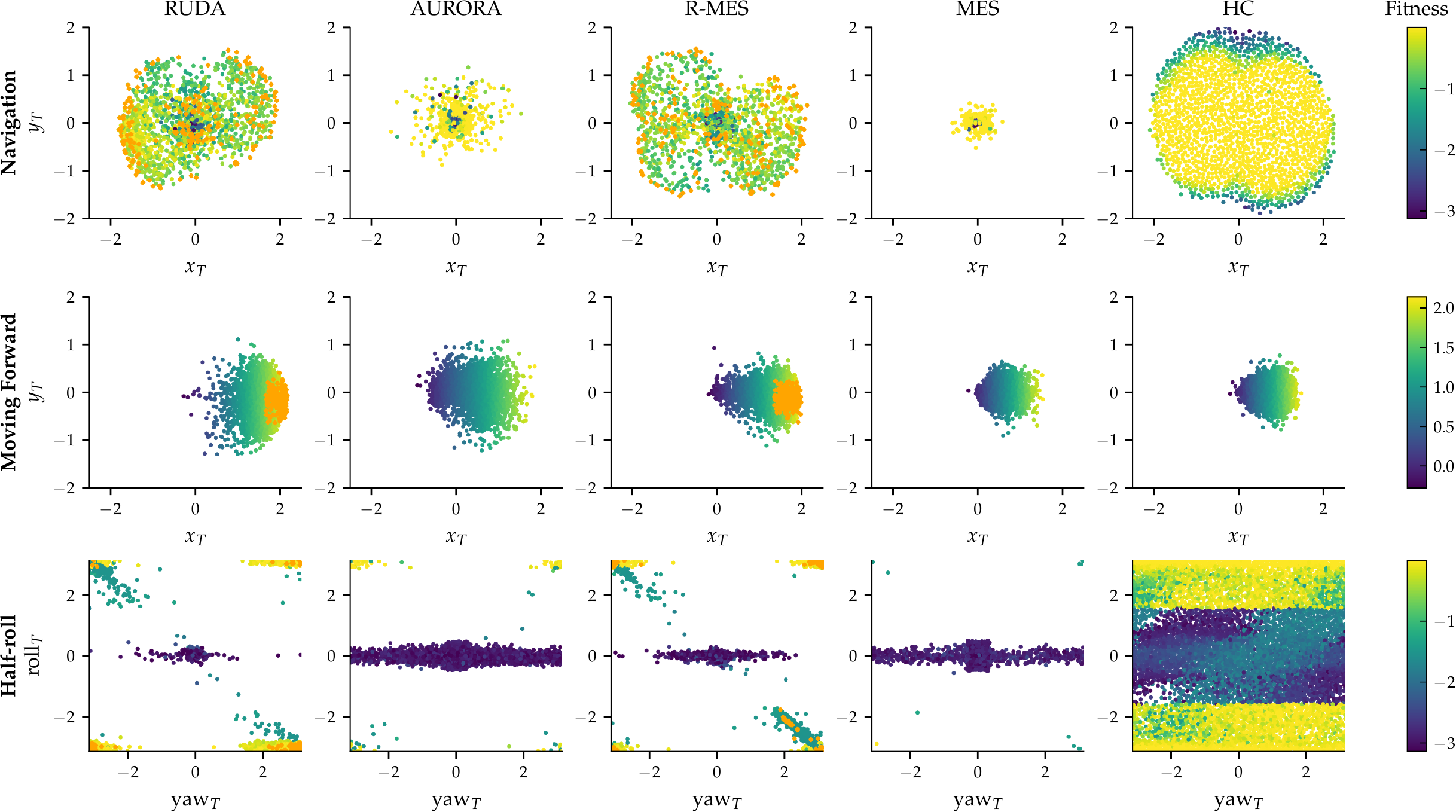}
    \caption{Containers obtained for each task and each algorithm variant.
    On each plot, each dot represents the BD obtained by one individual, and the colour represents the fitness score of that individual. 
    For relevance-based algorithms (\name{} and R-MeS), the orange dots represent the BDs of the individuals present in the buffer $\buffer$.
    }
    \label{fig:archives}
\end{figure*}

\subsection{Downstream Tasks and QD Tasks}

We make the distinction between downstream tasks and QD tasks.
A QD task evaluates the diversity and the performance of individuals present in the container returned by a QD algorithm.
A downstream task corresponds to a high-level problem that we intend to solve using the QD container.
%

%
%

For each experiment, we define a \textit{Task Solver}, a \textit{Task Score}, a \textit{Container Score} and an \textit{associated QD Task}.
The Task Solver chooses individuals from the container that are relevant to solve the task, and sends them to the buffer $\buffer$.
The Task Score characterises the overall performance of the Task Solver in solving the downstream task; contrary to the fitness score which evaluates the performance of an individual, the task score indicates the performance of a full container.
The Container Score is a theoretic measure of the expected performance, taking into account all individuals present in the container.
Finally, the associated QD Task is a set of BD and fitness function specifically hand-coded for succeeding at the downstream task; consequently, the QD Task is defined to be strongly correlated with the downstream task.


We consider three downstream tasks with the hexapod agent: Navigation, Moving Forward, and Half-roll.

\subsubsection{Navigation}

The Navigation task is inspired from (Chatzilygeroudis et al.) \citep{chatzilygeroudis2018reset} and (Kaushik et al.) \citep{kaushik2020adaptive}.
The container provided by the QD algorithm is used as a basis for choosing actions to reach successive goals, characterised by their $x,y$ position.

\textbf{Task Solver:} A goal is randomly positioned goal at a distance of approximately 5 meters from the starting position of the robot.
    The solver then iteratively chooses the individual from the repertoire that gets the closest to the goal, runs it for three seconds, and repeat the previous steps until the distance to the goal is lower than 5cm.
    
\textbf{Task Score:} We compute the number of actions required to reach a fixed number of 50 goals at specific positions, all separated by approximately 7 meters. 
The task score then corresponds to the negated average number of actions required to reach each separate goal.
    %
    
\textbf{Container Score:} 
    It corresponds to the coverage of the container in $x,y$ space. 
    That coverage is calculated by discretising the space of possible final positions $\left[-2.2m, 2.2m\right]^2$ into a $50\times 50$ grid.
    The coverage then corresponds the number of grid cells containing at least one individual from the unstructured container.

\textbf{Associated QD Task:} This task aims at finding a container of controllers reaching diverse final $(x_T, y_T)$ positions in $T=3$ seconds following circular trajectories. 
Hence the hand-coded BD in this case is $(x_T, y_T)$, while the fitness is the final orientation error as defined in the literature~\cite{Cully2013,chatzilygeroudis2018reset}.
%


\subsubsection{Moving Forward}

This task is inspired from the work of Cully et al. \cite{Cully2014robotsanimals}, which aims to obtain a container with a variety of ways to move forward at a high velocity.

\textbf{Task Solver:} it takes as input the entire container obtained from the AURORA phase, and returns the 10 individuals reaching the furthest position along the $x$ axis after 3 seconds. 
Those individuals are added to the buffer, with replacement.

\textbf{Task Score:} The score associated to that task corresponds to the \textit{maximal} position (along the $x$ axis) achieved by the container.

\textbf{Container Score:} corresponds to the \textit{mean} position (along the $x$ axis) of all individuals in the container.

\textbf{Associated QD Task:} The associated QD task follows the definition from Cully et al. \cite{Cully2014robotsanimals}, with the Duty Factor as hand-coded BD. 
The Duty Factor evaluates the proportion of time the each leg is in contact with the ground. 
%
%
The QD fitness promotes individuals achieving the highest $x$ position at the end of the episode.


\subsubsection{Half-roll}

The Half-roll task aims at obtaining a variety of ways to reach a position where the pitch angle of the robot is equal to $-\frac{\pi}{2}$ (i.e. where the robot falls on its back).

\textbf{Task Solver:} In this downstream task, the task solver returns the 10 individuals whose pitch angle is the closest to $-\frac{\pi}{2}$.

\textbf{Task Score:} The score associated to that task is the negated \textit{minimal} distance to a pitch angle of $-\frac{\pi}{2}$.

\textbf{Container Score:} The container score corresponds to the negated \textit{average} distance to a pitch angle of $-\frac{\pi}{2}$.

\textbf{Associated QD Task:} The QD task associated to the half-roll downstream task aims at finding a diversity of ways to perform half-rolls, i.e. behaviours such that the hexapod ends with its back on the floor.
In particular, the behaviours are characterised via the final yaw and roll angle of the torso.
And the fitness is the distance of the pitch angle $\alpha_{pitch}$ of the hexapod to $-\frac{\pi}{2}$.
\begin{equation}
    \vec{b}^{indiv} = \left(\alpha_{yaw}^{indiv}, \alpha_{roll}^{indiv}\right) \quad
    f^{indiv} = - \left\lvert \alpha_{pitch}^{indiv} - \left(-\frac{\pi}{2}\right) \right\rvert 
\end{equation}

\subsection{Compared algorithms and variants}

\begin{table}[t]
    \centering

    \begin{threeparttable}
    \begin{tabular}{l l c c}
    \toprule

Variant & BD & \begin{tabular}{@{}l@{}}
                   Uses Downstream \\
                   Task Solver \\
                 \end{tabular}
                 & Has Fitness \\

        \midrule
    	\addlinespace

\name{} & Unsupervised & \cmark & \xmark \\
AURORA & Unsupervised & \xmark & \cmark \\

\addlinespace

\relevantMeanStreams{} & Mean streams & \cmark & \xmark \\
\meanStreams{} & Mean streams & \xmark & \cmark \\

\addlinespace

HC & QD Task BD & \xmark & \cmark \\
    \bottomrule
    	
    \end{tabular}
    
    \end{threeparttable}

\caption[Summary]{Characteristics of the variants of \name{} under study, grouped based on their type of Behavioural Descriptor (BD).}
    \label{table:variants2}

\end{table}

We study two categories of variants: those that learn their BD autonomously, and those based on hand-defined BDs.
Those variants are summarised in Table~\ref{table:variants2}.

\subsubsection{Algorithms with Unsupervised Behavioural Descriptors}~ %

In addition to \name{}, we also study a version of \name{} that does not have a relevance-based mechanism, which corresponds to \aurora{}.

\textbf{\name{}}: 
 The dimensionality of the learnt BD space is set to 10.
 Similarly to prior work~\cite{Paolo2019taxons, grillotti2021unsupervised}, we suppose \name{} is completely QD-task-agnostic, which means that not only it automatically learns its BDs, but it also does not have access to the fitness function of the QD task $f^{indiv}$.
    
\textbf{\aurora{}}: 
It corresponds to \name{} without the relevance-based mechanism that updates the container metric.
    Contrary to \name{}, \aurora{} does not interact with a downstream task, but we consider \aurora{} has access to the fitness function $f^{indiv}$ of the QD task.
    This way, we will be able to evaluate the difference between using a fitness function, and using a relevance-based mechanism.

\subsubsection{Variants with Hand-defined Behavioural Descriptors}~ %

We consider three different variants having hand-defined BDs.
%
%

\textbf{Hand-Coded (HC)}: considers a low-dimensional BD that is defined by hand as the Behavioural Descriptor of the QD task (e.g. see Table~\ref{table:hc_qd_tasks}).
The HC variant corresponds to a standard QD algorithm with an unstructured archive~\cite{Cully2018QDFramework} and the mechanism of container size control introduced in previous work \citep{grillotti2021unsupervised}.

\textbf{R-MeS}: considers a Behavioural Descriptor with more information from the collected sensory than the Hand-Coded variant explained above.
    To calculate its BD, this variant considers the 18 data streams collected by the hexapod, and average each one of them to obtain a BD of dimension 18.
    This variant uses the same relevance-based bias as \name{}; and similarly to \name{}, it does not have access to the fitness function of the QD task $f^{indiv}$.
    The comparison between this variant and \name{} will be useful to evaluate the usefulness of the automatic dimensionality reduction algorithm.

\textbf{MeS}: presents the same characteristics as R-MeS, except that it does not use any relevance-based mechanism.
    As \aurora{}, MeS does not have access to the downstream task, but it uses the fitness function $f^{indiv}$.

\subsection{Implementation Details}

All our experiments were run for 15{,}000 iterations with a uniform QD selector.
We only use polynomial mutations as variation operators with $\eta_m = 10$, and a mutation rate of $0.3$.
The target container size $\targetscC$ of all algorithms is set to $5{,}000$ for the Moving Forward and the Half-roll tasks.
In the case of the Navigation Task, we set it to $1{,}500$ to obtain appropriate comparisons between the different approaches.
To keep the container size around $\targetscC$, we perform a container update every $T_{\cC} = 10$ iterations (as explained in Section~\ref{sec:background_qd}).
For relevance-based algorithms (\name{} and R-MeS), the downstream task execution period $\downstreamtaskperiod$ is set to 10 iterations, the maximal buffer size is set to $200$, and relevance scores (see Eq.~\ref{eq:relevance_score}) are calculated with $k=15$ nearest neighbours.

All our implementation is based on Sferes$_{\text{v2}}$ \cite{Mouret2010} 
and uses the DART simulator \cite{lee2018dart}, while the auto-encoders are coded and trained using the C++ API of PyTorch \citep{paszke2019pytorch}.

For each downstream task, each variant was run for 10 replications, and the statistical significance of the comparisons is measured using the Wilcoxon rank-sum test, with a Holm-Bonferroni correction \cite{holm1979simple}.
To facilitate the replications of the results, we stored pre-built experiments in a singularity container \cite{kurtzer2017singularity}, and made it available with the code at: \url{https://github.com/adaptive-intelligent-robotics/RUDA}. 

\section{Results}

\begin{figure}
    \centering
    \includegraphics[width=0.48\textwidth]{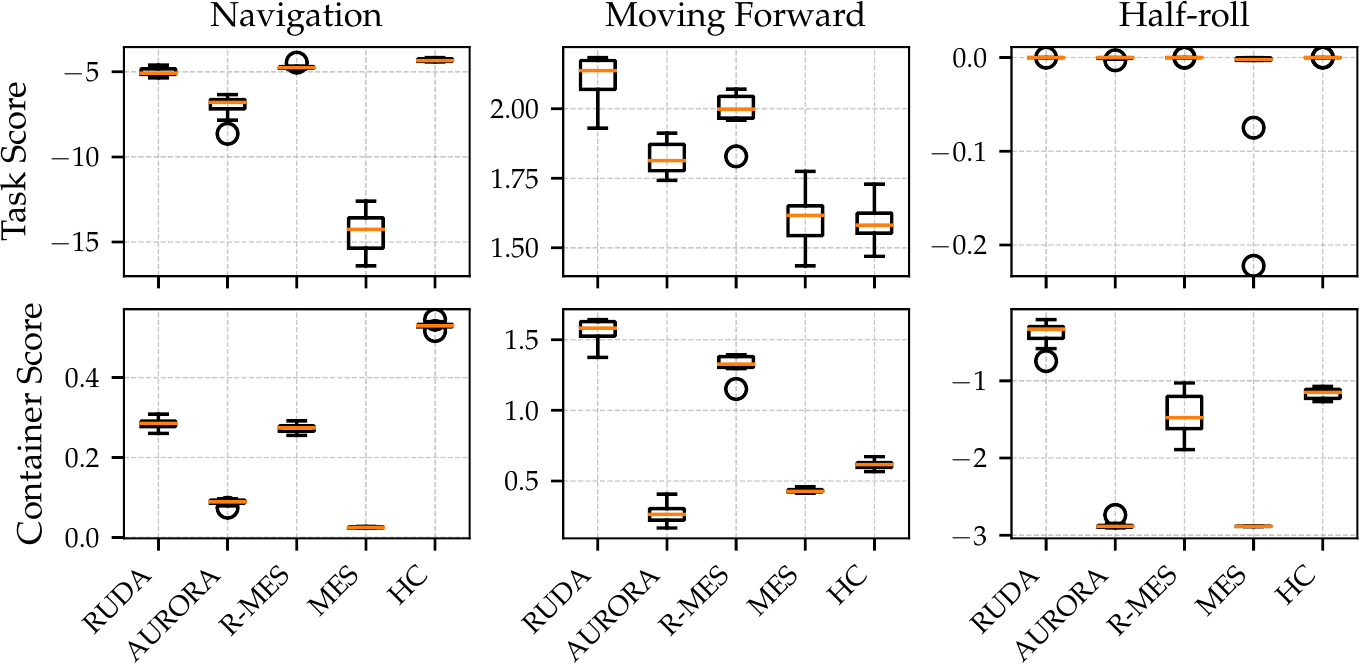}
    \caption{First row: Task Score, for the 3 tasks under study (Navigation, Moving Forward, Half-roll). The second row presents the Container Scores.}
    \label{fig:scores}
\end{figure}

\begin{figure*}
    \centering
    \includegraphics[width=0.75\textwidth]{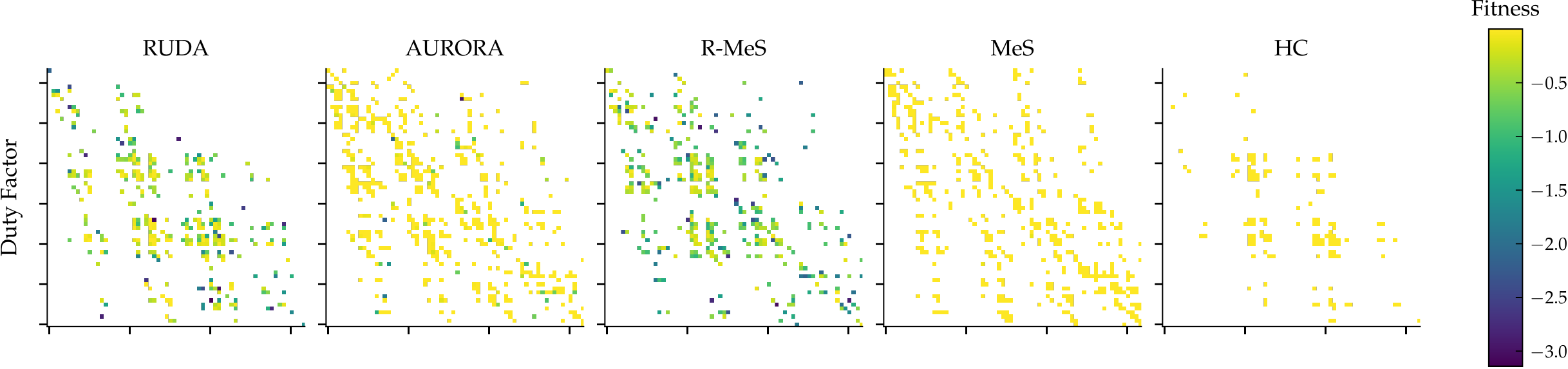}
    \caption{Visualisations of the obtained containers  from the "Navigation" task, for all algorithms under study.
    The containers are presented after having projected the individuals on the Duty Factor BD space.
    Each coloured pixel represents a grid cell of the BD space that has been filled with at least one individual.
    The colour is representative of the fitness.
    Note that the grids presented have six dimensions, they are presented using the same technique as in the work of Cully et al. \cite{Cully2014robotsanimals}.}
    \label{fig:04_diversity_analysis_omni}
\end{figure*}

\begin{figure*}
    \centering
    \includegraphics[width=0.75\textwidth]{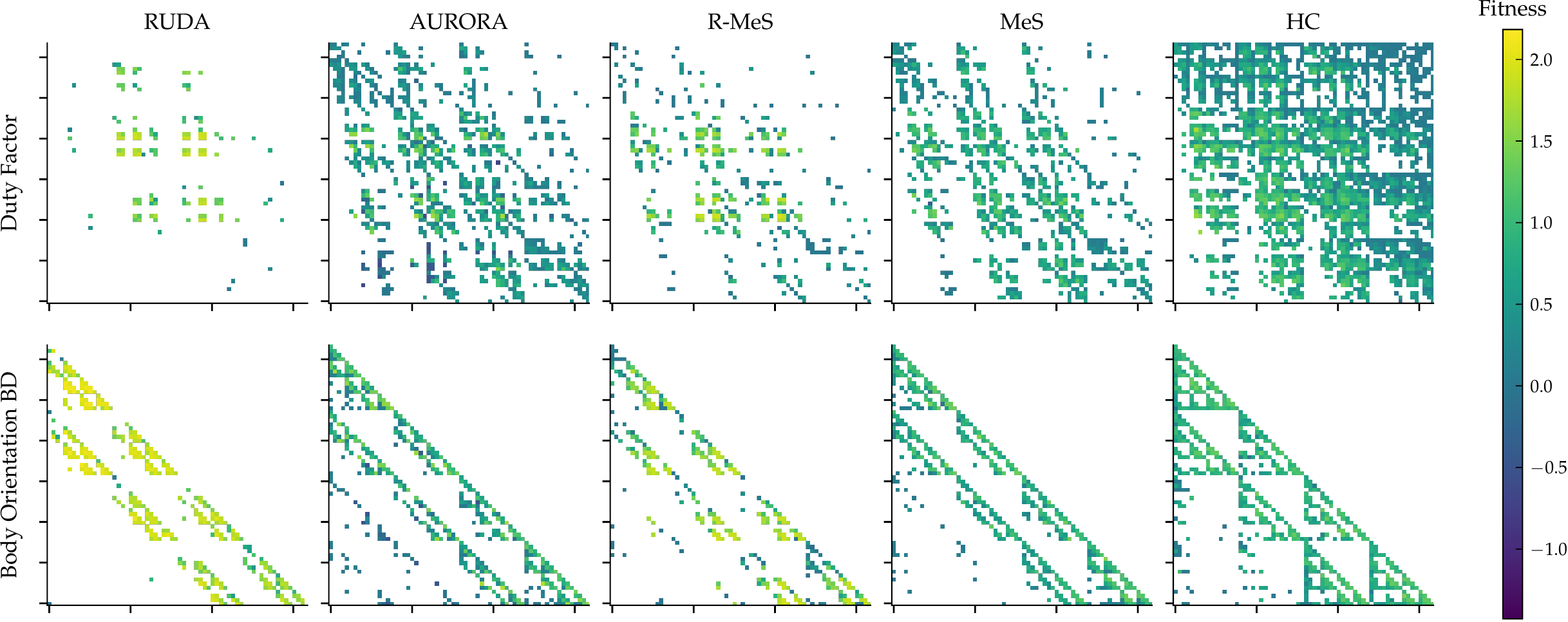}
    \caption{Visualisations of the obtained containers  from the "Moving Forward" task, for all algorithms under study.
    The first row presents the containers after having projected the individuals on the Duty Factor BD space.
    The second row shows the obtained containers after a projection on the Body Orientation BD space.
    }
    \label{fig:03_diversity_analysis_uni}
\end{figure*}

\begin{figure}
    \centering
    \includegraphics[width=\columnwidth]{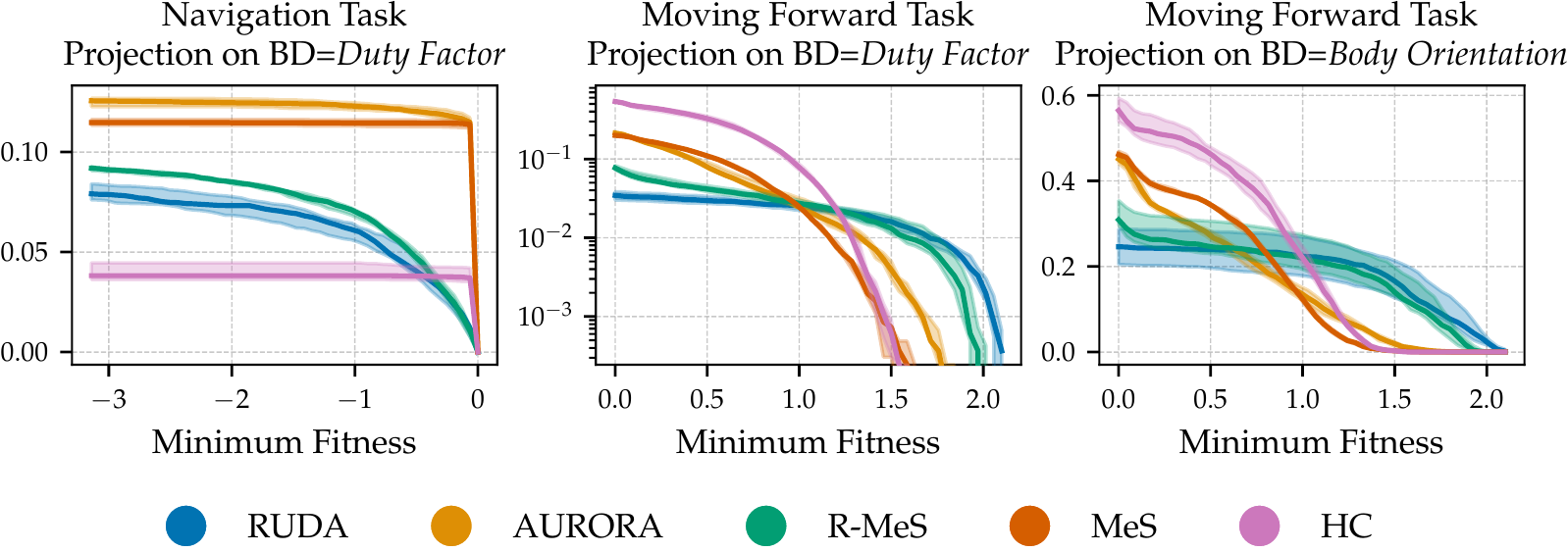}
    \caption{Coverage per minimum fitness, considered with different downstream tasks and projected in different BDs. 
    }
    \label{fig:05_coverage_per_fitness}
\end{figure}

We intend to show that the containers obtained through \name{} exhibit (1) relevance with respect to the downstream task under study, (2) while maintaining an overall behavioural diversity.

\subsection{Relevance to the Downstream Task}

We first aim at demonstrating that \name{} finds containers of individuals that are useful with respect to the task to solve.
The first thing to note is that, on the Moving Forward and Half-roll tasks, most of the container is situated near individuals present in the buffer (Fig.~\ref{fig:archives}).
    In those two tasks, this results in higher container scores for relevance-based mechanisms compared to \aurora{} and MeS (Fig.~\ref{fig:scores}-bottom, $p<2\times 10^{-3}$).
    
In particular, in the Moving Forward task, \name{} and R-MeS perform better than all other variants ($p<2\times 10^{-3}$).
    That is likely due to the shifting of the container distribution towards the most relevant individuals.
    Indeed, as most individuals in the container have a high $x$ position, then there is a high probability to generate individuals who also have a high $x$ position, and there are more chances to generate better offspring. 
    In other words, the QD selection is biased in the direction of the most useful individuals, which improves the results obtained by the relevance-based variants.

Also, in the Moving Forward and Half-roll tasks, \name{} also leads to higher container scores compared to the other relevance-based variant, R-MeS, (Fig.~\ref{fig:scores}-bottom, $p<5\times 10^{-3}$).
This shows that having a flexible encoding mechanism is useful for adapting the BD to maximise the relevance of the archive.
Moreover, in the tasks where the relevance is directly correlated with the fitness (Moving Forward and Half-roll), \name{} manages to find encodings that favour the most relevant individuals (see Fig.~\ref{fig:archives}).

In the case of the Navigation task, we can see that \name{} achieves significantly better task scores than \aurora{} (Fig.~\ref{fig:scores}, $p<2\times 10^{-3}$).
One possible explanation for this is that the containers obtained by \name{} present a wider coverage in terms of $x,y$ final positions than AURORA (Fig.~\ref{fig:scores}-bottom, and \ref{fig:archives}). 
That may also explain why using Hand-Coded BDs leads to the best scores.
We can also note that the coverage follows the peculiar distributions of the relevant individuals in the buffer: there is a higher-density of individuals executing either small manoeuvres or very large displacement. 
This makes sense as an effective way to solve this downstream task is to first walk as fast as possible in the direction of the goal and then adjust during the final approach. 
Hence, \name{} takes in average fewer actions to reach each goal, leading to a better task score.
%

    
    






\subsection{Container Diversity}

As explained before, the purpose of \name{} is not only to find specialised containers, but also containers that are diverse. This means that we expect \name{} to return a container of diverse relevant behaviours.
However, the notion of relevance is task-dependent, so we need to adjust the analysis to the task under study.
In particular, we take the containers returned by the different algorithms, and we project them in hand-coded BD spaces that are unrelated to the task.
Then we evaluate the achieved coverage in the BD space by considering the coverage for several minimum fitness scores.

\subsubsection{Coverage per minimum fitness} 

The containers returned by QD algorithms always result in a trade-off between the diversity of solutions and the total quality of the entire container.
To evaluate the coverage depending on the quality of the solutions, we study the evolution of the coverage given several minimum fitness scores.
With a fitness $f_{min}$ and a BD space discretised into a grid, the "coverage given minimum fitness $f_{min}$" is defined as: the percentage of cells with individuals whose fitness is higher than $f_{min}$ \cite{fontaine2021differentiable}.

\subsubsection{Navigation Task}

In the Navigation task, we have seen that \name{} and the Hand-Coded baseline return containers with diverse $x,y$ positions (Figs.~\ref{fig:archives} and~\ref{fig:scores}).
We intend to show that \name{} also discovers diverse ways to move to those positions.
To evaluate this gait diversity, we project the containers in the BD space "Duty Factors", and evaluate the coverage in that BD space.
Even when considering all individuals ($f_{min}=-3.5$), the container returned by \name{} on the Navigation task exhibits a higher coverage than the hand-coded container (Figs.~\ref{fig:04_diversity_analysis_omni} and~\ref{fig:05_coverage_per_fitness}, all $p$-values are $< 10^{-2}$).
The hand-coded variant does not have any incentive to promote diversity in the BD space of Duty Factors as it only attempts to maximise the diversity of final positions.
This may explain the poor performance of the Hand-coded variant in terms of diversity.

It is also worth noting that AURORA and MeS achieve a higher diversity than the relevance-based variants. 
That discrepancy is likely due to the fact that AURORA and MeS do not intend to specialise for any task.
On the contrary, as \name{} returns a container specialised for a downstream task, some diversity is inherently lost.

Finally, we see that the containers returned by the non-relevance based algorithms contain more high-performing individuals than \name{} and R-MeS. 
This phenomenon is due to the fact that \name{} does not optimise for that fitness function, which promotes circular trajectories.
Instead, here \name{} optimises for the downstream Navigation Task, and it is possible to succeed at that task while using individuals which are not performing circular trajectories.

\subsubsection{Moving Forward Task}

While \name{} produces containers of controllers to move forward, we also expect those containers to exhibit diversity in terms of other features.
We estimate this additional diversity by projecting the obtained containers in two classical BD spaces used for the Hexapod uni-directional task~\cite{Cully2014robotsanimals}: the Duty Factor, and the Body Orientation. The Duty Factor BD space is the same as defined previously, it is the one used by the hand-coded variant on the Moving Forward task.
The Body Orientation BD space characterises the amount of time spent by the torso of the hexapod at given orientations (more details can be found in the work of Cully et al. \cite{Cully2014robotsanimals}).

Figures~\ref{fig:03_diversity_analysis_uni} and~\ref{fig:05_coverage_per_fitness} show that the containers of \name{} present a lower coverage compared to the other variants that are not relevance-based ($p < 10^{-2} $).
However, when considering a minimum fitness score above $1.4$, which corresponds to a minimal $x_T$ of $1.4$ meters, the diversity exhibited by \name{} is better ($p < 2\times 10^{-3} $).
%
%
This confirms the fact that the container returned by \name{} not only specialises for a downstream task, but also still exhibits diversity in its specialisation.







\section{Conclusion and Future Work}

In this paper, we introduced \name{}, an extension of AURORA which aims to autonomously generate behavioural repertoires with a higher density of behaviours in the regions of the BD space that are relevant for the considered tasks.
Our experimental evaluation demonstrated that \name{} can adjust the distribution of behaviours depending on the situation, over the three considered tasks.

We also noticed that specialising for a downstream task often reduces the behavioural diversity, or moves the diversity towards the most relevant areas of the behavioural spaces.
It would be interesting to study whether all future methods to distort the learnt BD space will show the same influence on the trade-off between specialisation and diversity.
It would also be relevant to try our approach on more complex problems, such as multi-tasks problems~\cite{mouret2020quality}.

\begin{acks}
This work was supported by the Engineering and Physical Sciences Research Council (EPSRC) grant EP/V006673/1 project REcoVER. 
%
\end{acks}

\bibliographystyle{ACM-Reference-Format}
\bibliography{sample-base}


\begin{thebibliography}{41}


\ifx \showCODEN    \undefined \def \showCODEN     #1{\unskip}     \fi
\ifx \showDOI      \undefined \def \showDOI       #1{#1}\fi
\ifx \showISBNx    \undefined \def \showISBNx     #1{\unskip}     \fi
\ifx \showISBNxiii \undefined \def \showISBNxiii  #1{\unskip}     \fi
\ifx \showISSN     \undefined \def \showISSN      #1{\unskip}     \fi
\ifx \showLCCN     \undefined \def \showLCCN      #1{\unskip}     \fi
\ifx \shownote     \undefined \def \shownote      #1{#1}          \fi
\ifx \showarticletitle \undefined \def \showarticletitle #1{#1}   \fi
\ifx \showURL      \undefined \def \showURL       {\relax}        \fi
\providecommand\bibfield[2]{#2}
\providecommand\bibinfo[2]{#2}
\providecommand\natexlab[1]{#1}
\providecommand\showeprint[2][]{arXiv:#2}

\bibitem[\protect\citeauthoryear{Akkaya, Andrychowicz, Chociej, Litwin, McGrew,
  Petron, Paino, Plappert, Powell, Ribas, et~al\mbox{.}}{Akkaya
  et~al\mbox{.}}{2019}]%
        {akkaya2019solving}
\bibfield{author}{\bibinfo{person}{Ilge Akkaya}, \bibinfo{person}{Marcin
  Andrychowicz}, \bibinfo{person}{Maciek Chociej}, \bibinfo{person}{Mateusz
  Litwin}, \bibinfo{person}{Bob McGrew}, \bibinfo{person}{Arthur Petron},
  \bibinfo{person}{Alex Paino}, \bibinfo{person}{Matthias Plappert},
  \bibinfo{person}{Glenn Powell}, \bibinfo{person}{Raphael Ribas},
  {et~al\mbox{.}}} \bibinfo{year}{2019}\natexlab{}.
\newblock \showarticletitle{Solving rubik's cube with a robot hand}.
\newblock \bibinfo{journal}{\emph{arXiv preprint arXiv:1910.07113}}
  (\bibinfo{year}{2019}).
\newblock


\bibitem[\protect\citeauthoryear{Alvarez, Dahlskog, Font, and Togelius}{Alvarez
  et~al\mbox{.}}{2019}]%
        {alvarez2019empowering}
\bibfield{author}{\bibinfo{person}{Alberto Alvarez}, \bibinfo{person}{Steve
  Dahlskog}, \bibinfo{person}{Jose Font}, {and} \bibinfo{person}{Julian
  Togelius}.} \bibinfo{year}{2019}\natexlab{}.
\newblock \showarticletitle{Empowering quality diversity in dungeon design with
  interactive constrained map-elites}. In \bibinfo{booktitle}{\emph{2019 IEEE
  Conference on Games (CoG)}}. IEEE, \bibinfo{pages}{1--8}.
\newblock


\bibitem[\protect\citeauthoryear{Alvarez, Fernandez, Dahlskog, and
  Togelius}{Alvarez et~al\mbox{.}}{2020}]%
        {alvarez2020interactive}
\bibfield{author}{\bibinfo{person}{Alberto Alvarez}, \bibinfo{person}{Jose
  Maria Maria~Font Fernandez}, \bibinfo{person}{Steve Dahlskog}, {and}
  \bibinfo{person}{Julian Togelius}.} \bibinfo{year}{2020}\natexlab{}.
\newblock \showarticletitle{Interactive constrained map-elites: Analysis and
  evaluation of the expressiveness of the feature dimensions}.
\newblock \bibinfo{journal}{\emph{IEEE Transactions on Games}}
  (\bibinfo{year}{2020}).
\newblock


\bibitem[\protect\citeauthoryear{Bossens, Mouret, and Tarapore}{Bossens
  et~al\mbox{.}}{2020}]%
        {bossens2020learning}
\bibfield{author}{\bibinfo{person}{David~M Bossens},
  \bibinfo{person}{Jean-Baptiste Mouret}, {and} \bibinfo{person}{Danesh
  Tarapore}.} \bibinfo{year}{2020}\natexlab{}.
\newblock \showarticletitle{Learning behaviour-performance maps with
  meta-evolution}. In \bibinfo{booktitle}{\emph{Proceedings of the 2020 Genetic
  and Evolutionary Computation Conference}}. \bibinfo{pages}{49--57}.
\newblock


\bibitem[\protect\citeauthoryear{Cazenille}{Cazenille}{2021}]%
        {cazenille2021mcaurora}
\bibfield{author}{\bibinfo{person}{Leo Cazenille}.}
  \bibinfo{year}{2021}\natexlab{}.
\newblock \showarticletitle{Ensemble feature extraction for multi-container
  quality-diversity algorithms}. In \bibinfo{booktitle}{\emph{Proceedings of
  the Genetic and Evolutionary Computation Conference}}.
  \bibinfo{pages}{75--83}.
\newblock


\bibitem[\protect\citeauthoryear{Chatzilygeroudis, Cully, Vassiliades, and
  Mouret}{Chatzilygeroudis et~al\mbox{.}}{2021}]%
        {chatzilygeroudis2021quality}
\bibfield{author}{\bibinfo{person}{Konstantinos Chatzilygeroudis},
  \bibinfo{person}{Antoine Cully}, \bibinfo{person}{Vassilis Vassiliades},
  {and} \bibinfo{person}{Jean-Baptiste Mouret}.}
  \bibinfo{year}{2021}\natexlab{}.
\newblock \showarticletitle{Quality-Diversity Optimization: a novel branch of
  stochastic optimization}.
\newblock In \bibinfo{booktitle}{\emph{Black Box Optimization, Machine
  Learning, and No-Free Lunch Theorems}}. \bibinfo{publisher}{Springer},
  \bibinfo{pages}{109--135}.
\newblock


\bibitem[\protect\citeauthoryear{Chatzilygeroudis, Vassiliades, and
  Mouret}{Chatzilygeroudis et~al\mbox{.}}{2018}]%
        {chatzilygeroudis2018reset}
\bibfield{author}{\bibinfo{person}{Konstantinos Chatzilygeroudis},
  \bibinfo{person}{Vassilis Vassiliades}, {and} \bibinfo{person}{Jean-Baptiste
  Mouret}.} \bibinfo{year}{2018}\natexlab{}.
\newblock \showarticletitle{Reset-free trial-and-error learning for robot
  damage recovery}.
\newblock \bibinfo{journal}{\emph{Robotics and Autonomous Systems}}
  \bibinfo{volume}{100} (\bibinfo{year}{2018}), \bibinfo{pages}{236--250}.
\newblock


\bibitem[\protect\citeauthoryear{Coulom}{Coulom}{2007}]%
        {remi2006mcts}
\bibfield{author}{\bibinfo{person}{R{\'e}mi Coulom}.}
  \bibinfo{year}{2007}\natexlab{}.
\newblock \showarticletitle{Efficient Selectivity and Backup Operators in
  Monte-Carlo Tree Search}. In \bibinfo{booktitle}{\emph{Computers and Games}},
  \bibfield{editor}{\bibinfo{person}{H.~Jaap van~den Herik},
  \bibinfo{person}{Paolo Ciancarini}, {and} \bibinfo{person}{H.~H. L.
  M.~(Jeroen) Donkers}} (Eds.). \bibinfo{publisher}{Springer Berlin
  Heidelberg}, \bibinfo{address}{Berlin, Heidelberg}, \bibinfo{pages}{72--83}.
\newblock
\showISBNx{978-3-540-75538-8}


\bibitem[\protect\citeauthoryear{Cully}{Cully}{2019}]%
        {Cully2019}
\bibfield{author}{\bibinfo{person}{Antoine Cully}.}
  \bibinfo{year}{2019}\natexlab{}.
\newblock \showarticletitle{{Autonomous skill discovery with quality-diversity
  and unsupervised descriptors}}. In \bibinfo{booktitle}{\emph{GECCO 2019 -
  Proceedings of the 2019 Genetic and Evolutionary Computation Conference}}.
  \bibinfo{publisher}{Association for Computing Machinery, Inc},
  \bibinfo{pages}{81--89}.
\newblock
\showISBNx{9781450361118}
\showeprint[arxiv]{1905.11874}


\bibitem[\protect\citeauthoryear{Cully, Clune, Tarapore, and Mouret}{Cully
  et~al\mbox{.}}{2015}]%
        {Cully2014robotsanimals}
\bibfield{author}{\bibinfo{person}{Antoine Cully}, \bibinfo{person}{Jeff
  Clune}, \bibinfo{person}{Danesh Tarapore}, {and}
  \bibinfo{person}{Jean-Baptiste Mouret}.} \bibinfo{year}{2015}\natexlab{}.
\newblock \showarticletitle{Robots that can adapt like animals}.
\newblock \bibinfo{journal}{\emph{Nature}} \bibinfo{volume}{521},
  \bibinfo{number}{7553} (\bibinfo{year}{2015}), \bibinfo{pages}{503--507}.
\newblock


\bibitem[\protect\citeauthoryear{Cully and Demiris}{Cully and Demiris}{2018}]%
        {Cully2018QDFramework}
\bibfield{author}{\bibinfo{person}{Antoine Cully} {and}
  \bibinfo{person}{Yiannis Demiris}.} \bibinfo{year}{2018}\natexlab{}.
\newblock \showarticletitle{{Quality and Diversity Optimization: A Unifying
  Modular Framework}}.
\newblock \bibinfo{journal}{\emph{IEEE Transactions on Evolutionary
  Computation}} \bibinfo{volume}{22}, \bibinfo{number}{2} (\bibinfo{date}{apr}
  \bibinfo{year}{2018}), \bibinfo{pages}{245--259}.
\newblock
\showISSN{1089778X}
\showeprint[arxiv]{1708.09251}


\bibitem[\protect\citeauthoryear{Cully and Mouret}{Cully and Mouret}{2013}]%
        {Cully2013}
\bibfield{author}{\bibinfo{person}{Antoine Cully} {and}
  \bibinfo{person}{Jean~Baptiste Mouret}.} \bibinfo{year}{2013}\natexlab{}.
\newblock \showarticletitle{{Behavioral repertoire learning in robotics}}. In
  \bibinfo{booktitle}{\emph{GECCO 2013 - Proceedings of the 2013 Genetic and
  Evolutionary Computation Conference}}. \bibinfo{publisher}{ACM Press},
  \bibinfo{address}{New York, New York, USA}, \bibinfo{pages}{175--182}.
\newblock
\showISBNx{9781450319638}


\bibitem[\protect\citeauthoryear{Duarte, Gomes, Oliveira, and
  Christensen}{Duarte et~al\mbox{.}}{2016}]%
        {duarte2016evorbc}
\bibfield{author}{\bibinfo{person}{Miguel Duarte}, \bibinfo{person}{Jorge
  Gomes}, \bibinfo{person}{Sancho~Moura Oliveira}, {and}
  \bibinfo{person}{Anders~Lyhne Christensen}.} \bibinfo{year}{2016}\natexlab{}.
\newblock \showarticletitle{EvoRBC: evolutionary repertoire-based control for
  robots with arbitrary locomotion complexity}. In
  \bibinfo{booktitle}{\emph{Proceedings of the Genetic and Evolutionary
  Computation Conference 2016}}. ACM, \bibinfo{pages}{93--100}.
\newblock


\bibitem[\protect\citeauthoryear{Fontaine and Nikolaidis}{Fontaine and
  Nikolaidis}{2021}]%
        {fontaine2021differentiable}
\bibfield{author}{\bibinfo{person}{Matthew Fontaine} {and}
  \bibinfo{person}{Stefanos Nikolaidis}.} \bibinfo{year}{2021}\natexlab{}.
\newblock \showarticletitle{Differentiable Quality Diversity}.
\newblock \bibinfo{journal}{\emph{Advances in Neural Information Processing
  Systems}}  \bibinfo{volume}{34} (\bibinfo{year}{2021}).
\newblock


\bibitem[\protect\citeauthoryear{Fontaine, Lee, Soros, Silva, Togelius, and
  Hoover}{Fontaine et~al\mbox{.}}{2019}]%
        {fontaine2019}
\bibfield{author}{\bibinfo{person}{Matthew~C. Fontaine}, \bibinfo{person}{Scott
  Lee}, \bibinfo{person}{L.~B. Soros}, \bibinfo{person}{Fernando De~Mesentier
  Silva}, \bibinfo{person}{Julian Togelius}, {and} \bibinfo{person}{Amy~K.
  Hoover}.} \bibinfo{year}{2019}\natexlab{}.
\newblock \showarticletitle{Mapping Hearthstone Deck Spaces with Map-Elites
  with Sliding Boundaries}. In \bibinfo{booktitle}{\emph{Proceedings of The
  Genetic and Evolutionary Computation Conference}}. ACM.
\newblock


\bibitem[\protect\citeauthoryear{Fontaine, Togelius, Nikolaidis, and
  Hoover}{Fontaine et~al\mbox{.}}{2020}]%
        {fontaine2020covariance}
\bibfield{author}{\bibinfo{person}{Matthew~C Fontaine}, \bibinfo{person}{Julian
  Togelius}, \bibinfo{person}{Stefanos Nikolaidis}, {and}
  \bibinfo{person}{Amy~K Hoover}.} \bibinfo{year}{2020}\natexlab{}.
\newblock \showarticletitle{Covariance matrix adaptation for the rapid
  illumination of behavior space}. In \bibinfo{booktitle}{\emph{Proceedings of
  the 2020 genetic and evolutionary computation conference}}.
  \bibinfo{pages}{94--102}.
\newblock


\bibitem[\protect\citeauthoryear{Gravina, Liapis, and Yannakakis}{Gravina
  et~al\mbox{.}}{2016}]%
        {gravina2016surprisesearch}
\bibfield{author}{\bibinfo{person}{Daniele Gravina}, \bibinfo{person}{Antonios
  Liapis}, {and} \bibinfo{person}{Georgios Yannakakis}.}
  \bibinfo{year}{2016}\natexlab{}.
\newblock \showarticletitle{Surprise search: Beyond objectives and novelty}. In
  \bibinfo{booktitle}{\emph{Proceedings of the Genetic and Evolutionary
  Computation Conference 2016}}. \bibinfo{pages}{677--684}.
\newblock


\bibitem[\protect\citeauthoryear{Grillotti and Cully}{Grillotti and
  Cully}{2021}]%
        {grillotti2021unsupervised}
\bibfield{author}{\bibinfo{person}{Luca Grillotti} {and}
  \bibinfo{person}{Antoine Cully}.} \bibinfo{year}{2021}\natexlab{}.
\newblock \showarticletitle{Unsupervised Behaviour Discovery with
  Quality-Diversity Optimisation}.
\newblock \bibinfo{journal}{\emph{arXiv preprint arXiv:2106.05648}}
  (\bibinfo{year}{2021}).
\newblock


\bibitem[\protect\citeauthoryear{Holm}{Holm}{1979}]%
        {holm1979simple}
\bibfield{author}{\bibinfo{person}{Sture Holm}.}
  \bibinfo{year}{1979}\natexlab{}.
\newblock \showarticletitle{A simple sequentially rejective multiple test
  procedure}.
\newblock \bibinfo{journal}{\emph{Scandinavian journal of statistics}}
  (\bibinfo{year}{1979}), \bibinfo{pages}{65--70}.
\newblock


\bibitem[\protect\citeauthoryear{Hwangbo, Lee, Dosovitskiy, Bellicoso, Tsounis,
  Koltun, and Hutter}{Hwangbo et~al\mbox{.}}{2019}]%
        {hwangbo2019learning}
\bibfield{author}{\bibinfo{person}{Jemin Hwangbo}, \bibinfo{person}{Joonho
  Lee}, \bibinfo{person}{Alexey Dosovitskiy}, \bibinfo{person}{Dario
  Bellicoso}, \bibinfo{person}{Vassilios Tsounis}, \bibinfo{person}{Vladlen
  Koltun}, {and} \bibinfo{person}{Marco Hutter}.}
  \bibinfo{year}{2019}\natexlab{}.
\newblock \showarticletitle{Learning agile and dynamic motor skills for legged
  robots}.
\newblock \bibinfo{journal}{\emph{Science Robotics}} \bibinfo{volume}{4},
  \bibinfo{number}{26} (\bibinfo{year}{2019}), \bibinfo{pages}{eaau5872}.
\newblock


\bibitem[\protect\citeauthoryear{Jegorova, Doncieux, and Hospedales}{Jegorova
  et~al\mbox{.}}{2020}]%
        {jegorova2020behavioral}
\bibfield{author}{\bibinfo{person}{Marija Jegorova},
  \bibinfo{person}{St{\'e}phane Doncieux}, {and} \bibinfo{person}{Timothy~M
  Hospedales}.} \bibinfo{year}{2020}\natexlab{}.
\newblock \showarticletitle{Behavioral Repertoire via Generative Adversarial
  Policy Networks}.
\newblock \bibinfo{journal}{\emph{IEEE Transactions on Cognitive and
  Developmental Systems}} (\bibinfo{year}{2020}).
\newblock


\bibitem[\protect\citeauthoryear{Johns, Leutenegger, and Davison}{Johns
  et~al\mbox{.}}{2016}]%
        {johns2016deep}
\bibfield{author}{\bibinfo{person}{Edward Johns}, \bibinfo{person}{Stefan
  Leutenegger}, {and} \bibinfo{person}{Andrew~J Davison}.}
  \bibinfo{year}{2016}\natexlab{}.
\newblock \showarticletitle{Deep learning a grasp function for grasping under
  gripper pose uncertainty}. In \bibinfo{booktitle}{\emph{2016 IEEE/RSJ
  International Conference on Intelligent Robots and Systems (IROS)}}. IEEE,
  \bibinfo{pages}{4461--4468}.
\newblock


\bibitem[\protect\citeauthoryear{Kaushik, Desreumaux, and Mouret}{Kaushik
  et~al\mbox{.}}{2020}]%
        {kaushik2020adaptive}
\bibfield{author}{\bibinfo{person}{Rituraj Kaushik}, \bibinfo{person}{Pierre
  Desreumaux}, {and} \bibinfo{person}{Jean-Baptiste Mouret}.}
  \bibinfo{year}{2020}\natexlab{}.
\newblock \showarticletitle{Adaptive prior selection for repertoire-based
  online adaptation in robotics}.
\newblock \bibinfo{journal}{\emph{Frontiers in Robotics and AI}}
  (\bibinfo{year}{2020}), \bibinfo{pages}{151}.
\newblock


\bibitem[\protect\citeauthoryear{Kim, Coninx, and Doncieux}{Kim
  et~al\mbox{.}}{2021}]%
        {kim2021exploration}
\bibfield{author}{\bibinfo{person}{Seungsu Kim}, \bibinfo{person}{Alexandre
  Coninx}, {and} \bibinfo{person}{St{\'e}phane Doncieux}.}
  \bibinfo{year}{2021}\natexlab{}.
\newblock \showarticletitle{From exploration to control: learning object
  manipulation skills through novelty search and local adaptation}.
\newblock \bibinfo{journal}{\emph{Robotics and Autonomous Systems}}
  \bibinfo{volume}{136} (\bibinfo{year}{2021}), \bibinfo{pages}{103710}.
\newblock


\bibitem[\protect\citeauthoryear{Kingma and Ba}{Kingma and Ba}{2014}]%
        {kingma2014adam}
\bibfield{author}{\bibinfo{person}{Diederik~P Kingma} {and}
  \bibinfo{person}{Jimmy Ba}.} \bibinfo{year}{2014}\natexlab{}.
\newblock \showarticletitle{Adam: A method for stochastic optimization}.
\newblock \bibinfo{journal}{\emph{arXiv preprint arXiv:1412.6980}}
  (\bibinfo{year}{2014}).
\newblock


\bibitem[\protect\citeauthoryear{Kurtzer, Sochat, and Bauer}{Kurtzer
  et~al\mbox{.}}{2017}]%
        {kurtzer2017singularity}
\bibfield{author}{\bibinfo{person}{Gregory~M Kurtzer}, \bibinfo{person}{Vanessa
  Sochat}, {and} \bibinfo{person}{Michael~W Bauer}.}
  \bibinfo{year}{2017}\natexlab{}.
\newblock \showarticletitle{Singularity: Scientific containers for mobility of
  compute}.
\newblock \bibinfo{journal}{\emph{PloS one}} \bibinfo{volume}{12},
  \bibinfo{number}{5} (\bibinfo{year}{2017}), \bibinfo{pages}{e0177459}.
\newblock


\bibitem[\protect\citeauthoryear{Laversanne-Finot, P{\'e}r{\'e}, and
  Oudeyer}{Laversanne-Finot et~al\mbox{.}}{2021}]%
        {laversanne2021intrinsically}
\bibfield{author}{\bibinfo{person}{Adrien Laversanne-Finot},
  \bibinfo{person}{Alexandre P{\'e}r{\'e}}, {and} \bibinfo{person}{Pierre-Yves
  Oudeyer}.} \bibinfo{year}{2021}\natexlab{}.
\newblock \showarticletitle{Intrinsically motivated exploration of learned goal
  spaces}.
\newblock \bibinfo{journal}{\emph{Frontiers in neurorobotics}}
  (\bibinfo{year}{2021}), \bibinfo{pages}{109}.
\newblock


\bibitem[\protect\citeauthoryear{Lee, Grey, Ha, Kunz, Jain, Ye, Srinivasa,
  Stilman, and Liu}{Lee et~al\mbox{.}}{2018}]%
        {lee2018dart}
\bibfield{author}{\bibinfo{person}{Jeongseok Lee}, \bibinfo{person}{Michael~X
  Grey}, \bibinfo{person}{Sehoon Ha}, \bibinfo{person}{Tobias Kunz},
  \bibinfo{person}{Sumit Jain}, \bibinfo{person}{Yuting Ye},
  \bibinfo{person}{Siddhartha~S Srinivasa}, \bibinfo{person}{Mike Stilman},
  {and} \bibinfo{person}{C~Karen Liu}.} \bibinfo{year}{2018}\natexlab{}.
\newblock \showarticletitle{Dart: Dynamic animation and robotics toolkit}.
\newblock \bibinfo{journal}{\emph{Journal of Open Source Software}}
  \bibinfo{volume}{3}, \bibinfo{number}{22} (\bibinfo{year}{2018}),
  \bibinfo{pages}{500}.
\newblock


\bibitem[\protect\citeauthoryear{Lehman and Stanley}{Lehman and
  Stanley}{2011}]%
        {Lehman2011}
\bibfield{author}{\bibinfo{person}{Joel Lehman} {and}
  \bibinfo{person}{Kenneth~O Stanley}.} \bibinfo{year}{2011}\natexlab{}.
\newblock \showarticletitle{Abandoning objectives: Evolution through the search
  for novelty alone}.
\newblock \bibinfo{journal}{\emph{Evolutionary computation}}
  \bibinfo{volume}{19}, \bibinfo{number}{2} (\bibinfo{year}{2011}),
  \bibinfo{pages}{189--223}.
\newblock


\bibitem[\protect\citeauthoryear{Liapis, Mart{\'{i}}nez, Togelius, and
  Yannakakis}{Liapis et~al\mbox{.}}{2013}]%
        {Liapis2013}
\bibfield{author}{\bibinfo{person}{Antonios Liapis},
  \bibinfo{person}{H{\'{e}}ctor~P Mart{\'{i}}nez}, \bibinfo{person}{Julian
  Togelius}, {and} \bibinfo{person}{Georgios~N Yannakakis}.}
  \bibinfo{year}{2013}\natexlab{}.
\newblock \bibinfo{booktitle}{\emph{{Transforming Exploratory Creativity with
  DeLeNoX}}}.
\newblock \bibinfo{type}{{T}echnical {R}eport}.
\newblock


\bibitem[\protect\citeauthoryear{Masci, Meier, Cire{\c{s}}an, and
  Schmidhuber}{Masci et~al\mbox{.}}{2011}]%
        {masci2011stacked}
\bibfield{author}{\bibinfo{person}{Jonathan Masci}, \bibinfo{person}{Ueli
  Meier}, \bibinfo{person}{Dan Cire{\c{s}}an}, {and}
  \bibinfo{person}{J{\"u}rgen Schmidhuber}.} \bibinfo{year}{2011}\natexlab{}.
\newblock \showarticletitle{Stacked convolutional auto-encoders for
  hierarchical feature extraction}. In \bibinfo{booktitle}{\emph{International
  conference on artificial neural networks}}. Springer,
  \bibinfo{pages}{52--59}.
\newblock


\bibitem[\protect\citeauthoryear{Meyerson, Lehman, and Miikkulainen}{Meyerson
  et~al\mbox{.}}{2016}]%
        {meyerson2016learning}
\bibfield{author}{\bibinfo{person}{Elliot Meyerson}, \bibinfo{person}{Joel
  Lehman}, {and} \bibinfo{person}{Risto Miikkulainen}.}
  \bibinfo{year}{2016}\natexlab{}.
\newblock \showarticletitle{Learning behavior characterizations for novelty
  search}. In \bibinfo{booktitle}{\emph{Proceedings of the Genetic and
  Evolutionary Computation Conference 2016}}. \bibinfo{pages}{149--156}.
\newblock


\bibitem[\protect\citeauthoryear{Mouret and Clune}{Mouret and Clune}{2015}]%
        {Mouret2015}
\bibfield{author}{\bibinfo{person}{Jean-Baptiste Mouret} {and}
  \bibinfo{person}{Jeff Clune}.} \bibinfo{year}{2015}\natexlab{}.
\newblock \showarticletitle{{Illuminating search spaces by mapping elites}}.
\newblock  (\bibinfo{date}{apr} \bibinfo{year}{2015}).
\newblock
\showeprint[arxiv]{1504.04909}


\bibitem[\protect\citeauthoryear{Mouret and Doncieux}{Mouret and
  Doncieux}{2010}]%
        {Mouret2010}
\bibfield{author}{\bibinfo{person}{J.-B. Mouret} {and} \bibinfo{person}{S.
  Doncieux}.} \bibinfo{year}{2010}\natexlab{}.
\newblock \showarticletitle{{SFERES}v2: Evolvin' in the Multi-Core World}. In
  \bibinfo{booktitle}{\emph{Proc. of Congress on Evolutionary Computation
  (CEC)}}. \bibinfo{pages}{4079--4086}.
\newblock


\bibitem[\protect\citeauthoryear{Mouret and Maguire}{Mouret and
  Maguire}{2020}]%
        {mouret2020quality}
\bibfield{author}{\bibinfo{person}{Jean-Baptiste Mouret} {and}
  \bibinfo{person}{Glenn Maguire}.} \bibinfo{year}{2020}\natexlab{}.
\newblock \showarticletitle{Quality Diversity for Multi-task Optimization}.
\newblock  (\bibinfo{year}{2020}).
\newblock


\bibitem[\protect\citeauthoryear{Paolo, Coninx, Doncieux, and
  Laflaqui{\`e}re}{Paolo et~al\mbox{.}}{2021a}]%
        {paolo2021sparse}
\bibfield{author}{\bibinfo{person}{Giuseppe Paolo}, \bibinfo{person}{Alexandre
  Coninx}, \bibinfo{person}{St{\'e}phane Doncieux}, {and}
  \bibinfo{person}{Alban Laflaqui{\`e}re}.} \bibinfo{year}{2021}\natexlab{a}.
\newblock \showarticletitle{Sparse reward exploration via novelty search and
  emitters}. In \bibinfo{booktitle}{\emph{Proceedings of the Genetic and
  Evolutionary Computation Conference}}. \bibinfo{pages}{154--162}.
\newblock


\bibitem[\protect\citeauthoryear{Paolo, Coninx, Laflaqui{\`e}re, and
  Doncieux}{Paolo et~al\mbox{.}}{2021b}]%
        {paolo2021discovering}
\bibfield{author}{\bibinfo{person}{Giuseppe Paolo}, \bibinfo{person}{Alexandre
  Coninx}, \bibinfo{person}{Alban Laflaqui{\`e}re}, {and}
  \bibinfo{person}{Stephane Doncieux}.} \bibinfo{year}{2021}\natexlab{b}.
\newblock \showarticletitle{Discovering and Exploiting Sparse Rewards in a
  Learned Behavior Space}.
\newblock \bibinfo{journal}{\emph{arXiv preprint arXiv:2111.01919}}
  (\bibinfo{year}{2021}).
\newblock


\bibitem[\protect\citeauthoryear{Paolo, Laflaquiere, Coninx, and
  Doncieux}{Paolo et~al\mbox{.}}{2020}]%
        {Paolo2019taxons}
\bibfield{author}{\bibinfo{person}{Giuseppe Paolo}, \bibinfo{person}{Alban
  Laflaquiere}, \bibinfo{person}{Alexandre Coninx}, {and}
  \bibinfo{person}{Stephane Doncieux}.} \bibinfo{year}{2020}\natexlab{}.
\newblock \showarticletitle{Unsupervised learning and exploration of reachable
  outcome space}. In \bibinfo{booktitle}{\emph{2020 IEEE International
  Conference on Robotics and Automation (ICRA)}}. IEEE,
  \bibinfo{pages}{2379--2385}.
\newblock


\bibitem[\protect\citeauthoryear{Paszke, Gross, Massa, Lerer, Bradbury, Chanan,
  Killeen, Lin, Gimelshein, Antiga, et~al\mbox{.}}{Paszke
  et~al\mbox{.}}{2019}]%
        {paszke2019pytorch}
\bibfield{author}{\bibinfo{person}{Adam Paszke}, \bibinfo{person}{Sam Gross},
  \bibinfo{person}{Francisco Massa}, \bibinfo{person}{Adam Lerer},
  \bibinfo{person}{James Bradbury}, \bibinfo{person}{Gregory Chanan},
  \bibinfo{person}{Trevor Killeen}, \bibinfo{person}{Zeming Lin},
  \bibinfo{person}{Natalia Gimelshein}, \bibinfo{person}{Luca Antiga},
  {et~al\mbox{.}}} \bibinfo{year}{2019}\natexlab{}.
\newblock \showarticletitle{Pytorch: An imperative style, high-performance deep
  learning library}. In \bibinfo{booktitle}{\emph{Advances in neural
  information processing systems}}. \bibinfo{pages}{8026--8037}.
\newblock


\bibitem[\protect\citeauthoryear{P{\'{e}}r{\'{e}}, Forestier, Sigaud, and
  Oudeyer}{P{\'{e}}r{\'{e}} et~al\mbox{.}}{2018}]%
        {Pere2018}
\bibfield{author}{\bibinfo{person}{Alexandre P{\'{e}}r{\'{e}}},
  \bibinfo{person}{S{\'{e}}bastien Forestier}, \bibinfo{person}{Olivier
  Sigaud}, {and} \bibinfo{person}{Pierre-Yves Oudeyer}.}
  \bibinfo{year}{2018}\natexlab{}.
\newblock \showarticletitle{{Unsupervised Learning of Goal Spaces for
  Intrinsically Motivated Goal Exploration}}.
\newblock \bibinfo{journal}{\emph{6th International Conference on Learning
  Representations, ICLR 2018 - Conference Track Proceedings}}
  (\bibinfo{date}{mar} \bibinfo{year}{2018}).
\newblock
\showeprint[arxiv]{1803.00781}


\bibitem[\protect\citeauthoryear{Salehi, Coninx, and Doncieux}{Salehi
  et~al\mbox{.}}{2021}]%
        {salehi2021brns}
\bibfield{author}{\bibinfo{person}{Achkan Salehi}, \bibinfo{person}{Alexandre
  Coninx}, {and} \bibinfo{person}{Stephane Doncieux}.}
  \bibinfo{year}{2021}\natexlab{}.
\newblock \showarticletitle{BR-NS: an archive-less approach to novelty search}.
  In \bibinfo{booktitle}{\emph{Proceedings of the Genetic and Evolutionary
  Computation Conference}}. \bibinfo{pages}{172--179}.
\newblock


\end{thebibliography}

\appendix

\end{document}